\pdfoutput=1

\documentclass[11pt]{article}

\usepackage[final]{acl}

\usepackage{times}
\usepackage{latexsym}

\usepackage[T1]{fontenc}

\usepackage[utf8]{inputenc}

\usepackage{microtype}

\usepackage{inconsolata}

\usepackage{graphicx}

\usepackage{caption}
\usepackage{arydshln}
\usepackage{hyperref}
\usepackage{url}
\usepackage[export]{adjustbox}
\usepackage{multirow}
\newcommand{\quotes}[1]{``#1''}
\usepackage{booktabs}
\usepackage{amssymb}
\usepackage{pifont}
\usepackage{amsmath}
\usepackage{wrapfig}  
\usepackage{tcolorbox}
\newcommand\blfootnote[1]{
    \begingroup
    \renewcommand\thefootnote{}\footnote{#1}
    \addtocounter{footnote}{-1}
    \endgroup
}

\usepackage{colortbl}

\usepackage{enumitem}
\setlist[itemize]{noitemsep, topsep=3pt, partopsep=0pt, parsep=0pt}

\title{Length Representations in Large Language Models}

\author{Sangjun Moon$^{1\dagger}$, Dasom Choi$^{1\dagger}$, $^*$Jingun Kwon$^1$, \\ \textbf{Hidetaka Kamigaito}$^2$, \textbf{and Manabu Okumura}$^3$ \\
 $^1$Chungnam National University, $^2$Nara Institute of Science and Technology (NAIST) \\
 $^3$Institute of Science Tokyo \\
 {\tt \{202550430,somcandy08\}@o.cnu.ac.kr} \\
 {\tt jingun.kwon@cnu.ac.kr} \\
 {\tt kamigaito.h@is.naist.jp} \\
 {\tt oku@pi.titech.ac.jp}
 \\}

\begin{document}
\maketitle
\begin{abstract}
Large language models (LLMs) have shown remarkable capabilities across various tasks, that are learned from massive amounts of text-based data. Although LLMs can control output sequence length, particularly in instruction-based settings, the internal mechanisms behind this control have been unexplored yet. In this study, we provide empirical evidence on how output sequence length information is encoded within the internal representations in LLMs. In particular, our findings show that multi-head attention mechanisms are critical in determining output sequence length, which can be adjusted in a disentangled manner. By scaling specific hidden units within the model, we can control the output sequence length without losing the informativeness of the generated text, thereby indicating that length information is partially disentangled from semantic information. Moreover, some hidden units become increasingly active as prompts become more length-specific, thus reflecting the model's internal awareness of this attribute. Our findings suggest that LLMs have learned robust and adaptable internal mechanisms for controlling output length without any external control.
\blfootnote{$^*$ Corresponding author.}
\blfootnote{$^\dagger$ Equal contribution.}
\blfootnote{Our code is available at \url{https://github.com/Mcat00/gilab_length}.}
\end{abstract}

\section{Introduction}

Large language models (LLMs) have gained considerable attention in recent years for their remarkable task-solving capabilities~\citep{ouyang2022training,weifinetuned,bubeck2023sparksartificialgeneralintelligence}. LLMs are trained to predict the next token in a sequence. They can produce coherent and informative text, which demonstrates their implicit understanding of diverse linguistic structures~\citep{tenney-etal-2019-bert,niu-etal-2022-bert,begu2023largelinguisticmodelsanalyzing}. Furthermore, they also learn when to stop generating text to ensure that the output adheres to appropriate length constraints~\citep{juseon-do-etal-2024-instructcmp}. Controlling output sequence length in LLMs is crucial for real-world applications, such as text summarization~\citep{liu-etal-2018-controlling,makino-etal-2019-global,liu-etal-2022-length,kwon-etal-2023-abstractive}, machine translation~\citep{wu2016googlesneuralmachinetranslation,murray-chiang-2018-correcting,zhuocheng-etal-2023-addressing}, knowledge QA, and dialogue generation~\citep{10.1145/3366423.3380270,gupta-etal-2021-controlling}, that necessitate fitting content within specified length limits without losing informativeness. Therefore, the number of studies attempting to improve length controllability has increased drastically~\citep{shen-etal-2023-loose,jie-etal-2024-prompt,yuan2024followinglengthconstraintsinstructions}.

Based on advancements in instruction-based LLMs, it is observed that injecting constraints into prompts can further effectively control output length without requiring model modifications~\citep{juseon-do-etal-2024-instructcmp}. However, these prompt engineering methods mainly focus on external controls, and it has not been explored yet how LLMs internally encode and constrain output sequence length. 
Understanding these internal mechanisms is critical for achieving precise length control, while enhancing the interpretability and robustness of LLMs in generation-based systems.
Herein, we aim to investigate how output sequence length information is encoded within the internal representations of general transformer architectures. Specifically, we first investigate which components within LLM transformer layers contribute to length control. Our findings reveal that the outputs from multi-head attention mechanisms in the lower layers play a key role in determining and controlling output sequence length in a tunable and disentangled manner.

For this work, we utilize a sentence summarization task, which often requires adherence to desired summary lengths, and 
employ models from the Llama~\cite{llama32technicalreport}, Phi-3~\cite{abdin2024phi3technicalreporthighly}, and Qwen-2.5~\cite{qwen2025qwen25technicalreport} families.

We empirically demonstrate, based on human evaluations, that we can adjust output length during generation without losing the informativeness of texts by scaling specific hidden units within the outputs from the lower layers of multi-head attention mechanisms. For instance, multiplying certain hidden units with negative numbers results in longer text, while multiplying them by positive numbers generates more concise texts without losing informativeness. Furthermore, certain hidden units related to length information show increasing activity as prompts become more specific regarding length constraints. These units appear to be directly involved in controlling output length, indicating that LLMs have learned to process length-related information as a distinct feature, partially disentangled from other semantic information. Moreover, we find that the same highly activated hidden units are consistently involved in length control even after fine-tuning, regardless of length constraints in prompts~\citep{dai-etal-2023-gpt}.

\section{Related Work}

\noindent \textbf{Large Language Models.} In recent years, LLMs have achieved considerable success due to their remarkable task-solving abilities, specifically in zero-shot settings~\citep{Radford2019LanguageMA,NEURIPS2020_1457c0d6}. LLMs are broadly categorized into open and closed models. The open models, such as the Llama or Phi family, offer flexible access to modify their architectures, while the closed models, such as ChatGPT,\footnote{\url{https://chat.openai.com/}} have demonstrated remarkable reasoning abilities in various natural language processing tasks~\citep{jiao2023chatgptgoodtranslatoryes,peng-etal-2023-towards,laskar-etal-2023-systematic,ye-etal-2023-complementary,xie-etal-2023-empirical,xie-etal-2024-self}. Recent studies have focused on finding better methods to prompt LLMs~\citep{zhou2022teachingalgorithmicreasoningincontext,kojima2023largelanguagemodelszeroshot,zhou2023leasttomostpromptingenablescomplex}.

\noindent \textbf{Mechanistic Interpretability.}
Due to increasing interest in investigating the internal mechanisms of deep neural networks~\citep{ruker2023transparentaisurveyinterpreting}, significant attempts have been made to understand LLMs with a focus on models like BERT~\citep{tenney-etal-2019-bert,rogers-etal-2020-primer,niu-etal-2022-bert}, GPT~\citep{hanna2023doesgpt2computegreaterthan}, and even multimodal models~\citep{openaimultimodalneuron}. For instance, \cite{gurnee2024languagemodelsrepresentspace} showed that, when handling various prompts, LLMs learn linear representations of space and time across multiple scales, that show robustness. They also showed that next token prediction can be changed simply by disentangling hidden units related to time. \cite{heinzerling-inui-2024-monotonic} introduced directions that encode numeric properties in an interpretable manner; hence, by disentangling these representations, LLM prediction can change accordingly. 
There have been attempts to investigate how in-context learning with LLMs behaves similar to explicit fine-tuning for better understanding them~\citep{dai-etal-2023-gpt}. Early efforts to investigate how neural networks treat length information have focused on memory cell networks in LSTMs, as they recursively encode and decode sequences, though they failed to find single units related to length information~\citep{shi-etal-2016-neural}.

\noindent \textbf{Length Controllable Summarization.}
Text summarization aims to produce a concise summary from an original text by retaining informative contents~\citep{liu-etal-2018-controlling,takase-okazaki-2019-positional,Li_Zhu_Zhang_Zong_He_2020,he-etal-2022-ctrlsum}. As the summarization often requires additional constraints such as a desired summary length, previous studies have focused on learning length-specific parameters~\citep{kikuchi-etal-2016-controlling,schumann-etal-2020-discrete,ghalandari-etal-2022-efficient}, injecting direct constraints~\citep{takase-okazaki-2019-positional,makino-etal-2019-global}, or splitting the training dataset into specific length ranges~\citep{he-etal-2022-ctrlsum}. Recently, \cite{juseon-do-etal-2024-instructcmp} considered in-context learning and demonstrated that LLMs can control output sequence length through \quotes{length priming}. This method involves injecting more length-specific information into prompts, thereby allowing the model to adjust output sequence length without modifying model architectures or learning parameters. \cite{jie-etal-2024-prompt} considered length control types such as greater/smaller than a value with exhaustive model modifications by reinforcement learning.

To the best of our knowledge, this study is the first attempt to interpret how length information is encoded in LLMs and demonstrate how length-specific information is partially disentangled from semantic information. Furthermore, by comparing various length-specific prompts, we investigate how in-context learning and fine-tuning can influence the internal representations of LLMs. 
Finally, we demonstrate how disentangling length-specific hidden units can adjust output sequence length without losing informativeness.

\begin{table*}[h]

\begin{adjustbox}{width=1.8\columnwidth,center}
\centering
\Huge
\renewcommand{\arraystretch}{1.2}
\begin{tabular}{cl}
\toprule
\rowcolor{gray!10}
\textbf{Constraint} & \textbf{Instruction}\\
\midrule
No-constraint & Sentence:\{src\}The sentence without the less important tokens would be:\\
\midrule
Length & Sentence:\{src\}The sentence without the less important \{del\} tokens would be:\\
\midrule
\multirow{2}{*}{Priming}  & Sentence that consists of \{src len\} tokens:\{src\}The sentence that consists of\\ & \{keep\} tokens without the less important \{del\} tokens would be: \\
\bottomrule
\end{tabular}
\end{adjustbox}

\caption{Instruction formats. \quotes{src} indicates the placeholder for a source sentence, \quotes{del} denotes the placeholder for the number of deleted tokens, and \quotes{keep} and \quotes{src len} denote additional length information.}
\label{tab:instruction_format}
\end{table*}
\section{Finding Length Representations}
Our goal is to understand whether and how length representations are encoded in LLMs when using various length-constraint prompts. For this, we extracted outputs from different components and layers of transformer architectures during text generation. We then applied regression to predict the generation time steps from these hidden states. 

\noindent \textbf{Summarization Dataset.}
We used the \textbf{Google} sentence summarization dataset\footnote{\url{https://github.com/google-research-datasets/sentence-compression.git}}~\cite{filippova-altun-2013-overcoming} in an instruction-based format, following previous work\footnote{\url{https://github.com/JuseonDo/InstructCMP}}~\citep{juseon-do-etal-2024-instructcmp} because recent studies on length control still faces challenges in managing it with LLMs~\cite{jie-etal-2024-prompt,yuan2024followinglengthconstraintsinstructions}.
Table~\ref{tab:instruction_format} presents the instruction templates. 
As can be seen, in the \textbf{No-constraint} setting, the model summarizes a given sentence without considering a desired length, while in the \textbf{Length} setting, it summarizes the sentence with a specific desired length~\cite{fetahu-etal-2023-instructpts}. 
The \textbf{Priming} setting further considers more specific length information, such as the length of the given sentence and the number of tokens to keep~\citep{juseon-do-etal-2024-instructcmp}. Note that both \textbf{Length} and \textbf{Priming} are the state-of-the-art prompting methods for controlling output length in LLMs. 

The dataset includes 200k training, 1k validation, and 1k test pairs, where the average compression ratio in the test dataset is 0.45.
Length-specific prompts use ground-truth summary lengths.

\subsection{Models and Methods}
\noindent \textbf{Models.}
We performed our experiments using the Llama family of pre-trained instruction-based LLMs, which range from 1B to 70B parameters~\citep{touvron2023llama,grattafiori2024llama3herdmodels}, the Phi-3 family of mini~\citep{abdin2024phi3technicalreporthighly}, and the Qwen-2.5 family~\citep{qwen2025qwen25technicalreport}. 
Additionally, we considered how 4- and 8-bit quantizations influence length representations in LLMs. Furthermore, we fine-tuned an LLM on \textbf{Google} using QLoRA~\citep{dettmers2023qlora} to study how fine-tuning with length constraint prompts affect length-related internal representations. We followed previous work that incorporates such prompts to enhance output length control~\citep{juseon-do-etal-2024-instructcmp}.
We used greedy decoding in all experiments to eliminate randomness in generation.

\noindent \textbf{Gathering Model States.}
In the transformer, an input sentence $\mathbf{S}=\{s_1, s_2, \dots, s_n\}$ was first converted into vector embeddings, after which learned positional embeddings were added to form $\mathbf{S_{emb}}=\{s_1^e, s_2^e, \dots, s_n^e\}$. These embeddings were then normalized using layer normalization, expressed as $S_{\text{norm}} = \text{LN} \,(\mathbf{S_{\text{emb}}})$. Then, they were computed through query 
($\mathbf{W_Q}$), key 
($\mathbf{W_K}$), and value 
($\mathbf{W_V}$) 
matrices, and were fed into the transformer layers as follows:
\begin{align}
    & \text{MH}(Q, K, V) = \text{Concat}(\text{h}_1, \dots, \text{h}_h)W_O, \\
    & S_{\text{attn}} = \mathbf{S_{\text{emb}}} + \text{MH}(S^{Q}_{\text{norm}}, S^{K}_{\text{norm}}, S^{V}_{\text{norm}}), \\
    & S_{\text{ffn}} = \text{ReLU}(\text{LN}(S_{\text{attn}}) W_1 + b_1) W_2 + b_2, \\
    & S_{\text{out}} = S_{\text{attn}} + S_{\text{ffn}}, 
\end{align}
where each $\text{h}_i = \text{softmax}\left(\frac{QK^T}{\sqrt{d_k}}\right)V$ indicates a self-attention operation. 

We considered four outputs from the transformer layers: (1) multi-head attention, (2) multi-head attention combined with the original embeddings, (3) the outputs of feed-forward networks, and (4) an integration of (2) and (3). Each output represents a distinct level of encoded information derived from the original input sentence $\mathbf{S}$.
We conducted sentence summarization using prompts in three different settings: No-constraint, Length, and Priming. For each setting, we investigated these four outputs for each layer. During token generation, we saved each output with its corresponding numeric time step value, excluding the input token prompts~\cite{kaplan2024tokenswordsinnerlexicon}. For instance, we saved $n$ with its corresponding output when the model generated the $n$-th token. Appendix~\ref{appen:data} provides further details of data for predicting time steps from hidden states.

\begin{figure*}[ht!]
\centering
    \includegraphics[width=2\columnwidth]{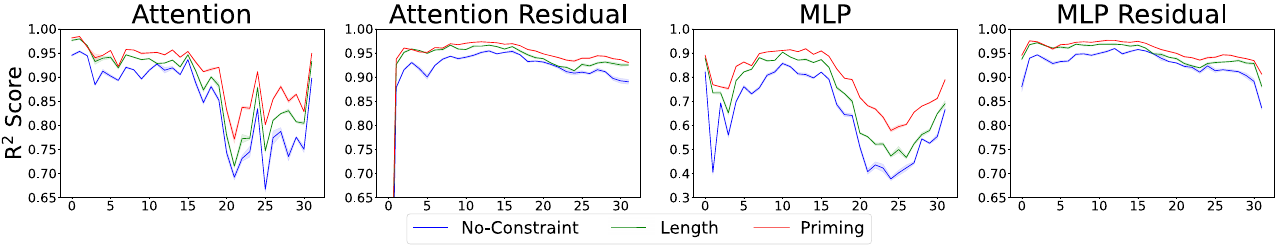}
    \caption{Average $R^2$ scores and standard errors with five runs for outputs of four different types of transformer layers using Llama-2-7B-Chat.}
    \label{fig:fig2}
    \vspace{-10pt}
\end{figure*}

\noindent \textbf{Neural Network Regression.}
To find evidence of length representations in LLMs, we applied a standard technique to predict a target label associated with labeled input data~\citep{shi-etal-2016-neural}, specifically, 
$\mathbf{X} \in \mathbb{R}^{m \times d_{\text{model}}}$, where $m$ refers to the number of data, $d_{\text{model}}$ is the dimensionality of a model's hidden states, and $\mathbf{Y}$ is a target that contains the generation time step as a numeric value for each corresponding $\mathbf{X}$. We used a two-layer neural network with a hidden layer of 100 neurons to predict $\mathbf{\hat{Y}}=\mathbf{W}_2(\text{ReLU}(\mathbf{W}_1\mathbf{X}+\mathbf{b}_1))+\mathbf{b}_2$. By investigating how well the model can predict the generation time step, we can gain insights into how length representations are encoded within the LLM's hidden states. 

To assess how well the time step can be predicted from its corresponding hidden state in LLMs, we considered the coefficient determination, $R^2$, as a standard regression metric to evaluate the overall performance. Appendix~\ref{sec:appendixA} provides details of hyper-parameters and settings.

\section{Length Representations in LLMs}
We explored which transformer layers and outputs contain length information, how length-specific prompts and quantization affect length representations, and the impact of fine-tuning on LLMs.

\noindent \textbf{Layer-wise Analysis for Length Representations.}
Figure~\ref{fig:fig2} shows the variation of $R^2$ for outputs from a transformer layer corresponding to Equations (1), (2), (3), and (4).
In the second layer, the outputs of Equation (1), which indicates the attention mechanism, show a stronger correlation with the length representations than the outputs from Equations (2), (3), and (4) for all prompts. 
Length representations decrease through LLM layers during token generation but increase in the final layer based on the attention outputs of Equation (1).
This indicates that the LLM captures length representations in the early stages, similar to how they capture semantic representations~\citep{niu-etal-2022-bert}. As such, the increase in length representations in the final layer indicates that the model may revisit this information to reinforce positional context.

\begin{table*}[h]
\renewcommand{\arraystretch}{1.0}
    \centering
    \small
\resizebox{\textwidth}{!}{
    \begin{tabular}{cccccccccccccc}
        \toprule
        \rowcolor{gray!10} &  & \multicolumn{12}{c}{\textbf{Layer type}} \\ \noalign{\vskip 0.1ex}\cline{3-14}
        \rowcolor{gray!10}
        & & \multicolumn{3}{c}{Attn Out} & \multicolumn{3}{c}{Attn Residual} & \multicolumn{3}{c}{MLP Out} & \multicolumn{3}{c}{MLP Residual} \\ \rowcolor{gray!10}
        \multirow{-3}{*}{\raisebox{0.5\height}{\textbf{Model}}} & \multirow{-3}{*}{\raisebox{0.5\height}{\textbf{Prom.}}} & F & S & \multicolumn{1}{c}{L} & F & S & \multicolumn{1}{c}{L} & F & S & \multicolumn{1}{c}{L} & F & S & \multicolumn{1}{c}{L} \\ \midrule 
        \multirow{3}{*}{\shortstack{Llama-2\\-7B}} & [1] & 0.94  & \textbf{0.95}  & 0.88  & 0.00  & 0.90  & 0.89  & 0.84  & 0.70  & 0.67  & 0.90  & 0.94  & 0.85  \\ 
        & [2] & 0.98  & \textbf{0.99}  & 0.93  & 0.11  & 0.94  & 0.93  & 0.89  & 0.77  & 0.70  & 0.95  & 0.97  & 0.89  \\ 
        & [3] & 0.98  & \textbf{0.99}  & 0.95  & 0.11  & 0.94  & 0.94  & 0.89  & 0.77  & 0.78  & 0.95  & 0.98  & 0.92  \\ \noalign{\vskip 0.5ex}\hline\noalign{\vskip 0.5ex}
        \multirow{3}{*}{\shortstack{Llama-2\\-13B}} & [1] & 0.95  & \textbf{0.96}  & 0.93  & 0.08  & 0.93  & 0.92  & 0.90  & 0.83  & 0.74  & 0.93  & 0.95  & 0.89  \\ 
        & [2] & \textbf{0.94}  & \textbf{0.94}  & 0.92  & 0.10  & 0.92  & 0.92  & 0.89  & 0.81  & 0.75  & 0.91  & \textbf{0.94}  & 0.91  \\ 
        & [3] & \textbf{0.99}  & \textbf{0.99}  & 0.91  & 0.17  & 0.96  & 0.92  & 0.92  & 0.81  & 0.72  & 0.97  & 0.98  & 0.89  \\ \noalign{\vskip 0.5ex}\hline\noalign{\vskip 0.5ex}
        \multirow{3}{*}{\shortstack{Llama-2\\-13B\\(finetuned)}} & [1] & \textbf{0.99}  & \textbf{0.99}  & 0.88  & 0.17  & 0.97  & 0.92  & 0.93  & 0.81  & 0.74  & 0.98  & 0.98  & 0.91  \\ 
        & [2] & \textbf{0.99} & \textbf{0.99}  & 0.87  & 0.21  & 0.97  & 0.93  & 0.92  & 0.83  & 0.78  & 0.98  & 0.98  & 0.91  \\ 
        & [3] & \textbf{0.99}  & \textbf{0.99}  & 0.90  & 0.16  & 0.96  & 0.93  & 0.92  & 0.85  & 0.83  & 0.97  & 0.98  & 0.92  \\ \noalign{\vskip 0.5ex}\hline\noalign{\vskip 0.5ex}
        \multirow{3}{*}{\shortstack{Llama-2\\-70B}} & [1] & 0.97  & \textbf{0.99}  & 0.95  & 0.16  & 0.93  & 0.92  & 0.83  & 0.81  & 0.82  & 0.95  & 0.98  & 0.92  \\ 
        & [2] & 0.97  & \textbf{0.99}  & 0.94  & 0.17  & 0.92 & 0.93  & 0.87  & 0.84  & 0.80  & 0.95  & 0.98  & 0.92  \\ 
        & [3] & \textbf{0.98}  & 0.97  & 0.91  & 0.18  & 0.91  & 0.89  & 0.82  & 0.76  & 0.78  & 0.94  & 0.95  & 0.88  \\ \noalign{\vskip 0.5ex}\hline\noalign{\vskip 0.5ex}
        \multirow{3}{*}{\shortstack{Llama-3\\-8B}} & [1] & 0.96 & \textbf{0.98}  & 0.91  & 0.20  & 0.86  & 0.91  & 0.70  & 0.74  & 0.78  & 0.88  & 0.95  & 0.88  \\ 
        & [2] & 0.96  & \textbf{0.97} & 0.93 & 0.16 & 0.88 & 0.93 & 0.72 & 0.75 & 0.79  & 0.90  & 0.96  & 0.89  \\ 
        & [3] & 0.97  & \textbf{0.98}  & 0.94 & 0.24  & 0.87  & 0.94  & 0.73  & 0.76  & 0.87  & 0.89  & 0.95  & 0.92  \\ \noalign{\vskip 0.5ex}\hline\noalign{\vskip 0.5ex}
        
        \multirow{3}{*}{\shortstack{Phi-3\\-mini\\-4k}} & [1] & 0.93  & \textbf{0.97}  & 0.91  & 0.07  & 0.80   & 0.91  & 0.61  & 0.66  & 0.55  & 0.84  & 0.95  & 0.86 \\
        & [2] & 0.94  & \textbf{0.97}  & 0.92 & 0.04 & 0.80  & 0.92  & 0.65  & 0.67  & 0.56  & 0.82  & 0.94  & 0.86  \\ 
        & [3] & 0.93  & \textbf{0.97}  & 0.89  & 0.07  & 0.77  & 0.90  & 0.48   & 0.63  & 0.58  & 0.80  & 0.95  & 0.84  \\ \noalign{\vskip 0.5ex}\hline\noalign{\vskip 0.5ex}
        \multirow{3}{*}{\shortstack{Phi-3.5\\-mini}} & [1] & 0.90 & \textbf{0.96} & 0.90 & 0.06 & 0.71 & 0.91 & 0.53 & 0.58 & 0.63 & 0.75 & 0.93 & 0.83  \\
        & [2] & 0.90 & \textbf{0.96} & 0.89 & 0.07 & 0.71 & 0.90 & 0.56 & 0.59 & 0.62 & 0.74 & 0.93 & 0.84 \\
        & [3] & 0.63 & \textbf{0.73} & 0.55 & 0.05 & 0.40 & 0.63 & 0.30 & 0.44 & 0.44 & 0.45 & \textbf{0.73} & 0.60 \\ \noalign{\vskip 0.5ex}\hline\noalign{\vskip 0.5ex}
        
        \multirow{3}{*}{\shortstack{Qwen-2.5\\-3B}} & [1] & 0.86 & \textbf{0.98} & 0.75 & 0.15 & 0.68 & 0.77 & 0.45 & 0.87 & 0.46 & 0.85 & 0.95 & 0.71  \\
        & [2] & 0.84 & \textbf{0.97} & 0.77 & 0.15 & 0.64 & 0.78 & 0.42 & 0.84 & 0.57 & 0.83 & 0.95 & 0.75 \\
        & [3] & 0.82 & \textbf{0.96} & 0.72 & 0.17 & 0.65 & 0.73 & 0.39 & 0.85 & 0.47 & 0.82 & 0.93 & 0.67 \\ \noalign{\vskip 0.5ex}\hline\noalign{\vskip 0.5ex}
        \multirow{3}{*}{\shortstack{Qwen-2.5\\-7B}} & [1] & 0.92 & \textbf{0.99} & 0.82 & 0.14 & 0.84 & 0.88 & 0.65 & 0.79 & 0.60 & 0.92 & 0.98 & 0.83  \\
        & [2] & 0.93 & \textbf{0.99} & 0.81 & 0.15 & 0.85 & 0.86 & 0.69 & 0.80 & 0.60 & 0.92 & 0.98 & 0.82 \\
        & [3] & 0.96 & \textbf{0.99} & 0.80 & 0.32 & 0.92 & 0.87 & 0.76 & 0.85 & 0.66 & 0.96 & 0.98 & 0.83 \\
        
        \bottomrule
    \end{tabular}
    }
 \caption{Average $R^2$ scores with five runs for different models with constraint prompt types from the first (F), second (S), and last (L) layers in LLMs. The standard errors are nearly zero. Prom. indicates the prompting method used, and [1], [2], and [3] indicate No-constraint, Length, and Priming prompts. }
  \label{tab:all_r2score}
  \vspace{-10pt}
\end{table*}

\begin{table*}[t]
    \centering
    \renewcommand{\arraystretch}{0.7}
    \Huge
    \resizebox{1\textwidth}{!}{
    \begin{tabular}{cccccccccccccc}
        \toprule
        \rowcolor{gray!10}
        &  & \multicolumn{12}{c}{\textbf{Layer type}} \\ 
        \cline{3-14}
        \rowcolor{gray!10}
        & & \multicolumn{3}{c}{Attn Out} & \multicolumn{3}{c}{Attn Residual} & \multicolumn{3}{c}{MLP Out} & \multicolumn{3}{c}{MLP Residual} \\
        \rowcolor{gray!10}
        \multirow{-3}{*}{\raisebox{0.5\height}{\textbf{Model}}} & \multirow{-3}{*}{\raisebox{0.5\height}{\textbf{Prom.}}} & First & Second & \multicolumn{1}{c}{Last} & First & Second & \multicolumn{1}{c}{Last} & First & Second & \multicolumn{1}{c}{Last} & First & Second & \multicolumn{1}{c}{Last} \\
        \midrule 
        \multirow{3}{*}{\shortstack{Llama-2 \\ -13B}} & [1] & 0.58/0.55 & 0.70/0.68 & \textbf{0.76}/\textbf{0.74} & 0.01/0.04 & 0.58/0.56 & 0.73/0.72 & 0.59/0.58 & 0.56/0.57 & 0.61/0.59 & 0.58/0.55 & 0.66/0.64 & 0.70/0.69 \\ 
        & [2] & \textbf{0.99}/\textbf{0.99} & \textbf{0.99}/\textbf{0.99} & 0.94/0.94 & 0.11/0.11 & 0.96/0.97 & 0.94/0.94 & 0.92/0.93 & 0.83/0.83 & 0.76/0.76 & 0.96/0.96 & 0.98/0.98 & 0.92/0.92 \\ 
        & [3] & \textbf{0.99}/\textbf{0.99} & \textbf{0.99}/\textbf{0.99} & 0.92/0.92 & 0.19/0.19 & 0.96/0.96 & 0.92/0.91 & 0.92/0.91 & 0.80/0.81 & 0.75/0.76 & 0.96/0.97 & 0.98/0.98 & 0.90/0.89 \\ \noalign{\vskip 0.5ex}\hline\noalign{\vskip 0.5ex}
        \multirow{3}{*}{\shortstack{Llama-2\\-13B\\(finetuned)}} & [1] & \textbf{0.99}/\textbf{0.99} & 0.98/0.98 & 0.87/0.87 & 0.19/0.22 & 0.96/0.97 & 0.92/0.92 & 0.91/0.92 & 0.80/0.80 & 0.75/0.77 & 0.97/0.97 & 0.99/0.98 & 0.90/0.90 \\ 
        & [2] & \textbf{0.99}/\textbf{0.99} & 0.98/\textbf{0.99} & 0.87/0.87 & 0.19/0.20 & 0.96/0.97 & 0.92/0.93 & 0.91/0.91 & 0.81/0.83 & 0.78/0.78 & 0.97/0.97 & 0.98/0.98 & 0.91/0.92 \\ 
        & [3] & \textbf{0.99}/\textbf{0.99} & \textbf{0.99}/0.98 & 0.90/0.90 & 0.19/0.19 & 0.95/0.96 & 0.93/0.93 & 0.91/0.92 & 0.86/0.86 & 0.82/0.82 & 0.96/0.97 & 0.98/0.98 & 0.92/0.92 \\ \noalign{\vskip 0.5ex}\hline\noalign{\vskip 0.5ex}
        \multirow{3}{*}{\shortstack{Llama-3\\-8B}} & [1] & 0.96/0.93 & \textbf{0.97}/\textbf{0.96} & 0.92/0.92 & 0.18/0.18 & 0.86/0.82 & 0.91/0.91 & 0.69/0.69 & 0.76/0.75 & 0.79/0.80 & 0.87/0.83 & 0.95/0.93 & 0.88/0.88 \\
        & [2] & 0.96/0.95 & \textbf{0.98}/\textbf{0.97} & 0.93/0.93 & 0.15/0.15 & 0.87/0.86 & 0.92/0.93 & 0.70/0.72 & 0.78/0.76 & 0.78/0.79 & 0.89/0.87 & 0.96/0.95 & 0.89/0.89 \\
        & [3] & 0.88/\textbf{0.86} & \textbf{0.91}/0.85 & 0.90/\textbf{0.86} & 0.16/0.14 & 0.79/0.64 & 0.84/0.82 & 0.64/0.53 & 0.60/0.55 & 0.59/0.61 & 0.79/0.70 & 0.87/0.79 & 0.75/0.73 \\ \noalign{\vskip 0.5ex}\hline\noalign{\vskip 0.5ex}
        
        \multirow{3}{*}{\shortstack{Phi-3\\-mini\\-4k}} & [1] & 0.94/0.94 & \textbf{0.97}/\textbf{0.97} & 0.92/0.92 & 0.07/0.07 & 0.80/0.82 & 0.93/0.93 & 0.61/0.64 & 0.66/0.67 & 0.52/0.53 & 0.84/0.84 & 0.95/0.96 & 0.87/0.87 \\
        & [2] & 0.94/0.94 & \textbf{0.97}/\textbf{0.97} & 0.93/0.93 & 0.06/0.05 & 0.82/0.82 & 0.93/0.92 & 0.61/0.63 & 0.67/0.65 & 0.53/0.52 & 0.84/0.83 & 0.95/0.96 & 0.87/0.86 \\
        & [3] & 0.92/0.93 & \textbf{0.97}/\textbf{0.98} & 0.90/0.90 & 0.10/0.12 & 0.74/0.75 & 0.90/0.90 & 0.48/0.46 & 0.61/0.66 & 0.58/0.60 & 0.78/0.78 & 0.94/0.96 & 0.84/0.85 \\ \noalign{\vskip 0.5ex}\hline\noalign{\vskip 0.5ex}

        \multirow{3}{*}{\shortstack{Qwen-2.5\\-1.5B}} & [1] & 0.81/0.82 & \textbf{0.96}/\textbf{0.96} & 0.51/0.53 & 0.12/0.13 & 0.67/0.69 & 0.82/0.83 & 0.55/0.59 & 0.69/0.67 & 0.63/0.65 & 0.81/0.81 & 0.93/0.93 & 0.77/0.75 \\
        & [2] & 0.83/0.82 & \textbf{0.96}/\textbf{0.97} & 0.54/0.56 & 0.12/0.12 & 0.66/0.66 & 0.87/0.86 & 0.58/0.58 & 0.71/0.69 & 0.64/0.61 & 0.82/0.81 & 0.95/0.95 & 0.80/0.77 \\
        & [3] & 0.67/0.80 & \textbf{0.92}/\textbf{0.97} & 0.23/0.38 & 0.10/0.15 & 0.54/0.65 & 0.55/0.85 & 0.50/0.52 & 0.56/0.65 & 0.41/0.50 & 0.66/0.78 & 0.82/0.94 & 0.49/0.77 \\
        \bottomrule
    \end{tabular}
    }
    \caption{Average $R^2$ scores with 8-bit and full-precision settings based on five runs. In each cell, x/y represents the 8-bit quantization and full-precision scores. The standard errors are nearly zero.}
    \label{tab:all_r2score8bit}
\end{table*}

\noindent \textbf{Influence of Length-specific Prompts.}
Table~\ref{tab:all_r2score} shows the results of $R^2$ for outputs, which include Llama and Phi LLMs with a 4-bit quantization setting. The results reveal that the attention output consistently has higher $R^2$ scores than the other outputs, particularly in the second layer for the Llama- and Phi-3 families, regardless of model sizes.
However, we observed a notable decrease in performance in the first layer, particularly in the attention residual. This indicates that the initial input sequence embeddings do not effectively contain length information; however, these representations progressively accumulate it through the layers. Although the length-specific prompting method (Priming) can precisely control output sequence length~\citep{juseon-do-etal-2024-instructcmp}, it does not increase the $R^2$ when using all hidden units for prediction. However, when we fine-tuned the models, we found that for every model, regardless of the prompts used, the $R^2$ scores were improved.

\noindent \textbf{Quantization on Length Representations.}
Table~\ref{tab:all_r2score8bit} shows the results with 8-bit and full-precision settings. The results are similar to those obtained with 4-bit quantization, wherein length representations are more prominently encoded in the attention outputs from the second layer than the other outputs. This indicates that whether 4- or 8-bit quantization is applied does not significantly affect the LLMs' capabilities to encode length representations. Therefore, the attention mechanism of the second layer consistently captures length representations across different precision levels even for different models with varying sizes.

\section{Disentangling Length Representations}
The previous section explored which components and layers contain length representations for output sequence length with varying prompts. While the second attention layer has a strong correlation with length representations, this does not indicate which hidden units are actually responsible for controlling the output sequence length. 
Thus, the specific hidden units must be identified for a better understanding of LLMs' length control.

\noindent \textbf{Do Length-specific Prompts Affect Inner Length Representations?}
We additionally trained separate neural network regression models on each single hidden unit from the second layer of the attention outputs in Llama-2-13B-Chat, which has a total of 5,120 hidden units.
Table~\ref{tab:eachhidden} shows the results for hidden units of the top-5 highest $R^2$ scores.
Compared to the No-constraint and Length prompting methods, length-related hidden units become more active in representing length information when we used more length-specific prompts Priming. 
In the zero-shot setting, No-constraint and Length prompts share similar top-5 hidden units for the length representation, while the Priming prompt activates different units, showing a shift in length capture and stronger activation in top-\textit{k} units.
After fine-tuning, the hidden units for the length representation became nearly identical across prompting methods, because the model learned the precise length control. Interestingly, the same top-3 hidden units are activated with the Priming prompt in the zero-shot and fine-tuning settings. 
This finding shows that specific length-related units consistently activate during Priming, thus guiding LLMs in output length control and revealing in-context learning as implicit fine-tuning~\citep{dai-etal-2023-gpt}.

\begin{table}[h]
    \centering
    \renewcommand{\arraystretch}{0.5}
    \Huge
    \resizebox{0.5\textwidth}{!}{
    \begin{tabular}{cccccccc}
    \rowcolor{gray!10}
        \toprule
        Setting & Prompting & 1$^{st}$ & 2$^{nd}$ & 3$^{rd}$ & 4$^{th}$ & 5$^{th}$ & Avg 30 \\
        \midrule 
        \multirow{6}{*}{Zero-shot} 
        & \multirow{2}{*}{No-constraint} & 0.11 & 0.10 & 0.09 & 0.07 & 0.07 & \multirow{2}{*}{0.06} \\
        & & (2,100) & (110) & (435) & (3,499) & (190) &  \\
        \noalign{\vskip 0.5ex}\cdashline{3-8}\noalign{\vskip 0.5ex} 
        & \multirow{2}{*}{Length} & 0.14 & 0.10 & 0.06 & 0.05 & 0.05 & \multirow{2}{*}{0.05} \\
        & & (2,100) & (110) & (435) & (321) & (1,411) &  \\
        \noalign{\vskip 0.5ex}\cdashline{3-8}\noalign{\vskip 0.5ex} 
        & \multirow{2}{*}{Priming} & \textbf{0.38} & \textbf{0.32} & \textbf{0.23} & \textbf{0.19} & \textbf{0.18} & \multirow{2}{*}{\textbf{0.08}} \\
        & & (371) & (2,741) & (1,380) & (4,698) & (4,554) &  \\
        \noalign{\vskip 0.5ex}\hline\noalign{\vskip 0.5ex} 
        \multirow{6}{*}{\shortstack{Fine-tuning}} 
        & \multirow{2}{*}{No-constraint} & \textbf{0.42} & 0.35 & \textbf{0.34} & 0.28 & \textbf{0.26} & \multirow{2}{*}{0.19} \\
        & & (2,741) & (1,380) & (371) & (4,698) & (2,282) & \\
        \noalign{\vskip 0.5ex}\cdashline{3-8}\noalign{\vskip 0.5ex} 
        & \multirow{2}{*}{Length} & 0.39 & 0.38 & 0.37 & 0.28 & 0.25 & \multirow{2}{*}{0.20} \\
        & & (1,380) & (371) & (2,741) & (4,372) & (4,698) & \\
        \noalign{\vskip 0.5ex}\cdashline{3-8}\noalign{\vskip 0.5ex} 
        & \multirow{2}{*}{Priming} & 0.40 & \textbf{0.39} & \textbf{0.34} & \textbf{0.31} & \textbf{0.26} & \multirow{2}{*}{\textbf{0.21}} \\
        & & (371) & (2,741) & (1,380) & (4,372) & (1,419) & \\
        \bottomrule
    \end{tabular}
    }    
    \caption{$R^2$ scores for individual hidden unit. 
    The numbers in parentheses indicate an index of hidden units from the second layer of the attention mechanisms.}
    \label{tab:eachhidden}
\end{table}

\begin{figure*}[h]
\centering
    \includegraphics[width=2\columnwidth]{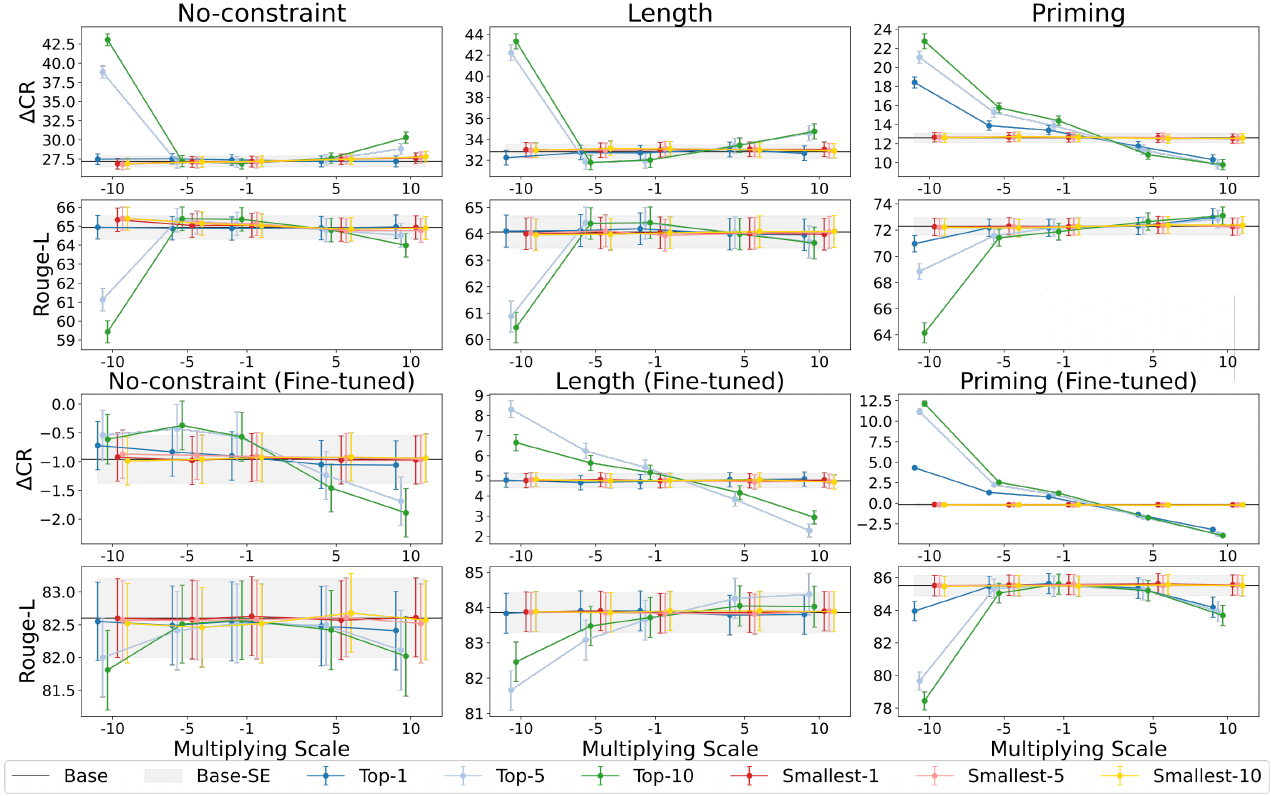}

    \caption{$\Delta$CR and Rouge-L scores change with standard errors by multiplying scale in the Llama-2-13B-Chat. (a) and (b) mean zero-shot and fine-tuning settings, respectively. Base means original scores without scale modification (i.e., the multiplying scale is 1). The gray color represents the standard errors of the Base.}
    \label{fig:zero}
\end{figure*}

\begin{figure*}[h]
\centering
    \includegraphics[width=2\columnwidth]{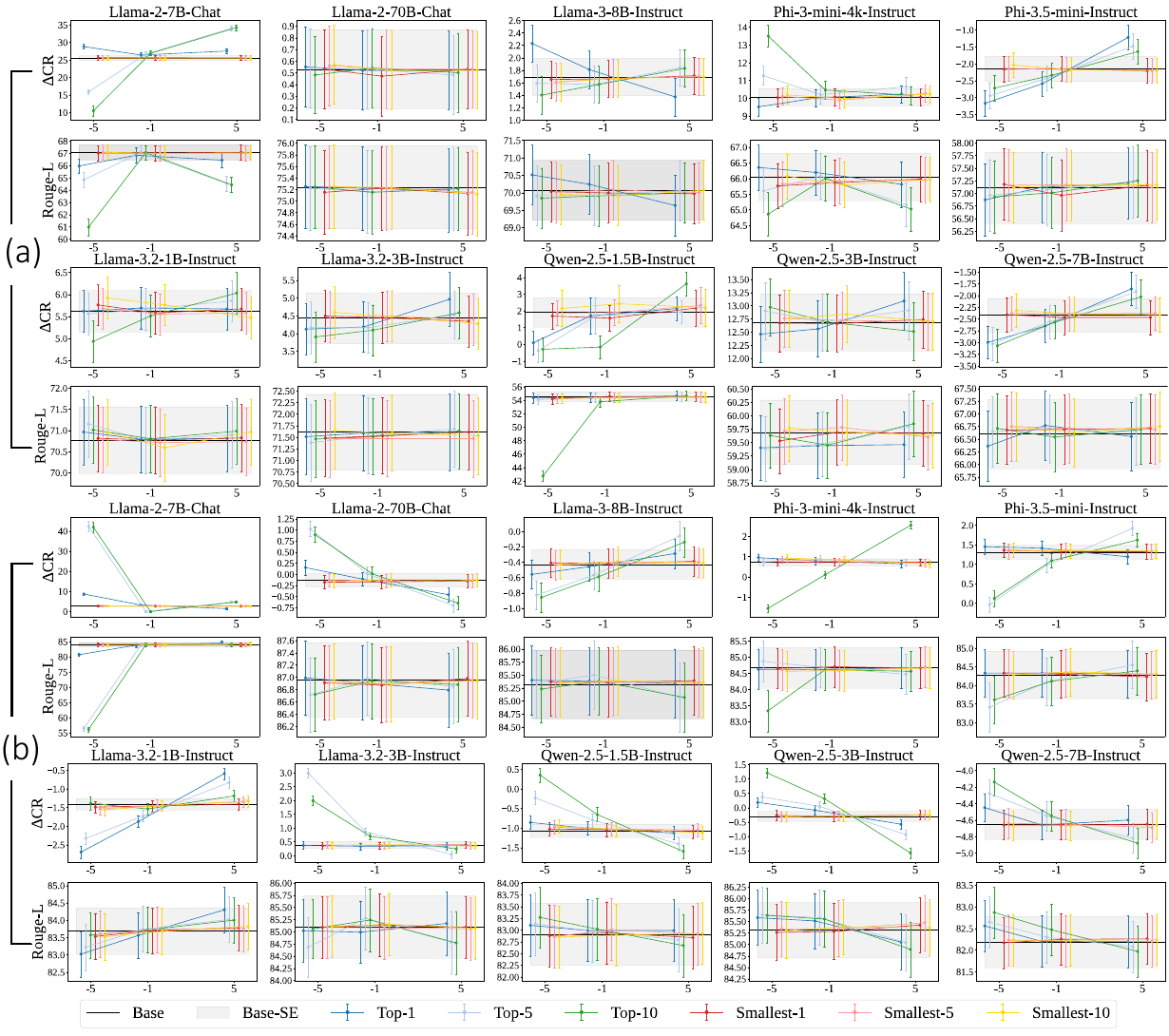}
    \caption{Results in (a) zero-shot settings and (b) fine-tuning settings with the Priming prompt.}
    \label{fig:zero2}
\end{figure*}

\noindent \textbf{Does Scaling Length Representations Affect Model-Generated Text?}
Since identified length-related units do not guarantee their actual involvement in length representations within LLMs~\cite{10.1162/tacl_a_00519,10.1162/coli_a_00422}, we investigate the effect of scaling these representations on model-generated text. Specifically, we disentangled the top-\textit{k} and smallest-\textit{k} activated hidden units in the second layer's multi-head attention by scaling them with positive or negative values. The scaling was applied to all output token positions except the input prompts.\footnote{We also explored other scaling methods, such as applying scaling only to the first output token, but found they were ineffective at controlling output lengths.} This approach demonstrates that the identified units contribute to length representations in LLMs and they are partially disentangled from semantic representations.

\noindent \textbf{Evaluation Metrics.}
We used Rouge-L (R-L)~\citep{lin-2004-rouge} to evaluate the informativeness of the summarized sentences.
To evaluate length control performance, we used $\Delta$CR, the arithmetical difference between model-generated and gold compression ratios. The compression ratio is the number of summary tokens divided by the number of source tokens. A $\Delta$CR close to zero indicates that the generated summaries have a compression ratio similar to the gold summaries, with higher values indicating longer summaries and lower values indicating shorter ones. Thus, deviations in $\Delta$CR from zero often lead to lower R-L scores due to reduced alignment with the gold summary~\cite{makino-etal-2019-global}.

\noindent \textbf{Results in Zero-Shot Settings.}
Figure~\ref{fig:zero} (a) presents the results of applying negative or positive scaling factors to the top-\textit{k} and smallest-\textit{k} activated hidden units in zero-shot settings. When using more length-specific prompts of Priming, we observed more consistent changes of $\Delta$CR with modifying only the top-1 hidden unit. We think this is because Priming contains more highly activated hidden units related to length representations than No-constraint and Length. Thus, the highly activated length-related units provide better length representations and length controllability, as shown in Table~\ref{tab:eachhidden}. Additionally, the output sequence length changes according to increases in the scaling factor. While multiplying positive and negative values enables the LLM to produce shorter and longer summaries, respectively, than the original hidden units, particularly in the Priming prompt, in the No-constraint and Length prompts, the LLM does not generate shorter summaries even when positive scaling values were applied. 

As for R-L scores, disentangling the top-\textit{k} units improves performance, particularly in the Priming prompt due to improved length alignment with the gold summary. This finding indicates that adjusting the most highly activated length-related units not only controls length but also enhances the informativeness of the generated text. However, when we applied large scaling factors, such as -10 or 10, the R-L scores slightly decrease when the No-constraint and Length prompts were used. In comparison, for Priming, which is more length-specific prompts, continues to improve performance even when we applied a large scaling factor of 10.
Disentangling the smallest-\textit{k} units does not lead to significant changes in output sequence length, thus indicating that these units are less involved in encoding length information. For selected smallest-\textit{k} units, the individual $R^2$ scores are nearly 0. 

\noindent \textbf{Results in Fine-Tuning Settings.}
Figure~\ref{fig:zero} (b) shows the results of applying negative or positive scaling factors to the top-\textit{k} and smallest-\textit{k} activated hidden units in fine-tuning settings. In contrast to the previous zero-shot settings, we obtained more stable results for all prompts when we disentangled the hidden units. While multiplying positive scaling values results in generating shorter summaries, multiplying negative values produces longer summaries. This is because fine-tuning has strengthened the LLM's reliance on the top-\textit{k} length-related units for precise length control. 
While large scaling factors lead to greater changes in $\Delta$CR and R-L for the Priming prompt, higher overall R-L scores are maintained. The decrease in R-L occurs because the Priming prompt in fine-tuning settings already achieves precise control, resulting in a $\Delta$CR close to zero. Applying large scaling factors causes deviations from zero and leads to a decline in R-L scores.
Disentangling the smallest-\textit{k} has minimal impact on sequence length among all prompts. Specifically, there are no significant changes in output sequence length when the smallest-\textit{k} hidden units were modified. 

\noindent \textbf{Results Using Different LLMs.}
Figure~\ref{fig:zero2} shows the results using different LLMs. While Llama-2-7B-Chat and Llama-2-13B-Chat used standard multi-head attentions during their pre-training steps, other LLMs employed grouped-query attentions~\cite{grattafiori2024llama3herdmodels,abdin2024phi3technicalreporthighly}. When we scaled the top-\textit{k} hidden units by multiplying scaling factors in the LLMs which employed the multi-head attentions, we observed variations in output length under zero-shot settings while scaling the smallest-\textit{k} hidden units did not impact length control during generation. In contrast, scaling the top-\textit{k} hidden units in the LLMs that employed the grouped-query attentions did not effectively control output length. However, after fine-tuning these LLMs, we found that disentangling the top-\textit{k} hidden units effectively controls output length. Additional experimental results using beam and top-\textit{k} sampling decoding strategies are in Appendix~\ref{appen:other_decoding}.

We also experimented with machine translation and story generation tasks using WMT16~\cite{bojar-etal-2016-findings} and ROCStories~\cite{mostafazadeh-etal-2016-corpus} test datasets. We disentangled identified length-related units from the summarization prompts. Figure~\ref{fig:Trans_result} and ~\ref{fig:roc_result} show the results. We observed that length-specific units are globally shared regardless of tasks. The details are in Appendix~\ref{appen:trans}.

\subsection{Human Evaluation and Case Study}

\textbf{Human Evaluation.} We conducted human evaluations to further assess the effect of disentangling length-related units. 
Note that we separately evaluated the zero-shot and fine-tuning settings; thus, their scales might be different. We sampled 100 instances for each setting from the Google test dataset. Using Amazon Mechanical Turk, we assigned a total of 80 evaluators who held both US high school and bachelor's degrees for grading the results, with scores from 1 to 5 (5 is the best), in terms of conciseness (Conc) and informativeness (Info). 

Table~\ref{tab:human} shows the results.
In the zero-shot and fine-tuning settings, adjusting the length-related hidden units with positive scaling factors generally enhances conciseness but slightly decreases informativeness because of 
generating shorter summaries. In some cases, generated summaries are already short~\cite{juseon-do-etal-2024-instructcmp}, and so conciseness scores are slightly higher when positive scaling was applied than those from the Base scale 1.
In contrast, negative scaling improves informativeness but slightly decreases conciseness due to the production of longer summaries, which can be an inherent trade-off between conciseness and informativeness when controlling output sequence length in summarization~\citep{kikuchi-etal-2016-controlling,makino-etal-2019-global}.

\begin{table}
    \begin{adjustbox}{width=0.7\columnwidth,center}
    \Huge
    \centering
    \begin{tabular}{ccccc}
    \rowcolor{gray!10}
        \toprule
        &  \multicolumn{2}{c}{Zero-shot} & \multicolumn{2}{c}{Fine-tuning} \\ 
        
        \cmidrule(lr){2-3} \cmidrule(lr){4-5}  \rowcolor{gray!10}
        Scale  &  Conc. & Infor. & Conc. & Infor. \\
        \midrule
        -10  &  3.56  &  \textbf{3.71}$^\dagger$   & \underline{3.33}  &  \textbf{3.34}$^\dagger$ \\
        1     &  3.59  &  3.70   &  3.46 &  3.31  \\
        Gold  &  3.58  &  3.68   &  3.45  &  3.28  \\
        10    &  3.59  &  \underline{3.63}   &  \textbf{3.47}$^\dagger$  &  \underline{3.19} \\
        \bottomrule
    \end{tabular}
    \end{adjustbox}
    \caption{The results of human evaluations using the Priming prompt with Llama-2-13B-Chat.$\dagger$ indicates the improvement for scales between 10 and -10 is significant using paired-bootstrap-resampling with 100,000 random samples (p$<$0.05)~\protect\citep{koehn}.} %
    \label{tab:human}
\end{table}

\noindent \textbf{Case Study.}
We conducted a detailed case study to analyze the effects of disentangling length-related hidden units by comparing the generated outputs for different scaling factors with the source and the gold summary. Table~\ref{tab:example} and~\ref{tab:case2} present examples. We observed changes in the generated summaries based on different scaling factors. In particular, when negative scaling was applied, the generated summaries became longer than the Base summary by incorporating redundant information from the source. In comparison, applying positive scaling values leads to shorter summaries by focusing on important content similar to the gold summary. When we disentangled the smallest-\textit{k} hidden units, the generated summaries remained unchanged, regardless of the scaling factors, consistently producing the same summary as the Base.

\begin{table*}
    \renewcommand{\arraystretch}{1}
    \centering
    \small
    \resizebox{0.9\textwidth}{!}{
    \begin{tabular}{p{0.25\columnwidth}p{0.2\columnwidth}p{1.05\columnwidth}p{0.2\columnwidth}}
        \toprule
        \rowcolor{gray!10}
        \multicolumn{2}{c}{\raisebox{-0.8\height}{\textbf{Type}}} & \raisebox{-0.8\height}{\textbf{Text}} & \textbf{Length (\#word)} \\
        \midrule
        \multicolumn{2}{c}{\multirow{3}{*}{Source}} & Armenian national's midfielder Aras Ozbiliz may miss the friendly match against Russia, technical director Vardan Minasyan told reporters ahead of the match. & 22 \\
        \multicolumn{2}{c}{Gold}   & Aras Ozbiliz may miss the friendly match against Russia. & 9 \\
        \midrule
        \multirow{4}{*}{\textbf{Top-10}}&
        Scale 5  & Armenian midfielder may miss \textcolor{blue}{Russia} \textcolor{gray}{against} \textcolor{blue}{match.} & 6 (-1) \\
        & Scale 10 & Armenian midfielder may miss \textcolor{blue}{Russia} \textcolor{gray}{against} \textcolor{blue}{match.} & 6 (-1) \\
        & Scale -5 & Armenian midfielder \textcolor{red}{Aras Ozbiliz} may miss \textcolor{red}{the} match against Russia. & 10 (+3) \\
        & Scale -10 & \textcolor{blue}{Aras Ozbiliz} may miss \textcolor{red}{the friendly} match against Russia, \textcolor{red}{technical director Vardan Minasyan told reporters ahead of the match}. & 19 (+12) \\
        \midrule
        \multirow{1}{*}{\textbf{Base (Scale 1)}} &
         & Armenian midfielder may miss match against Russia. & 7 \\
        \midrule
        \multirow{4}{*}{\textbf{Smallest -10}} &
        Scale 5  & Armenian midfielder may miss match against Russia. & 7 \\
        &Scale 10 & Armenian midfielder may miss match against Russia. & 7 \\
        &Scale -5 & Armenian midfielder may miss match against Russia. & 7 \\
        &Scale -10 & Armenian midfielder may miss match against Russia. & 7 \\
        \bottomrule
    \end{tabular}
    }
    \caption{Summarization example by scaling with Llama-2-13B-Chat in zero-shot Priming. The highlighted part represents the changed part from the Base text. The gray and red tokens indicate deleted and added tokens, respectively, while the blue token represents tokens that have changed their positions.}
    \label{tab:example}
\end{table*}

\section{Discussion and Conclusion}
We examined how LLMs encode output sequence length in their internal representations.
Our findings empirically demonstrated that the outputs from the second layer's attention mechanisms showed a strong correlation with the generation time step, thus indicating that length representations were captured early in the process. We also found that this pattern is consistent with different models with different sizes 
and continues to be robust even when 4- and 8-bit quantizations were applied. 
Furthermore, we analyzed individual hidden unit from the second layer attention outputs and found that certain hidden units are highly activated and directly contribute to the process of representing length information. Moreover, these units became more active when length-specific prompts such as Priming were used. This finding indicates that LLMs adjust their internal representations based on the input prompts. Furthermore, by scaling these length-related hidden units, we effectively controlled the output sequence length without losing informativeness, 
that indicates that length information is partially disentangled from semantic representations within LLMs. 
Finally, our results revealed that fine-tuning further improves the LLMs' capabilities by reinforcing reliance on the top-\textit{k} length-related units. We also found the same activation of specific hidden units in the Priming prompt are shared between zero-shot and fine-tuning settings, that indicates LLMs have constructed robust internal mechanisms for controlling output sequence length, and in-context learning performs similarly to implicit fine-tuning~\citep{dai-etal-2023-gpt}. 

\section*{Limitations}
Our findings have important implications for the interpretability and controllability of LLMs in natural language generation tasks. Understanding how length information is internally encoded allows for more precise length control over generated outputs, which is crucial in applications, such as summarization and machine translation, where adhering to length constraints is often required. We focused on a summarization dataset because summarization is a widely used task for controlling output length~\cite{kikuchi-etal-2016-controlling,takase-okazaki-2019-positional}, and recent studies on the summarization task still face challenges when using LLMs for length control~\cite{juseon-do-etal-2024-instructcmp,yuan2024followinglengthconstraintsinstructions,wang-etal-2024-positionid}. Thus, we considered other tasks such as machine translation and story generation and discussed them in Appendix~\ref{appen:trans}.

While we used neural networks to identify length representations in LLMs, there are inherent limitations, particularly due to their ability to decode functionally irrelevant information from model representations~\cite{10.1162/tacl_a_00519,10.1162/coli_a_00422}. To validate that the identified hidden units are involved in length control in LLMs, we disentangled their top-\textit{k} and smallest-\textit{k} length-related units.
However, our findings face a limitation in their application to LLMs that use grouped-query attentions in zero-shot settings. Despite disentangling the top-\textit{k} units, the proposed methods do not effectively control or influence the model's internal representations for length control. Moreover, whether our findings can extend to models employing Mixture of Experts (MoE) architectures or other types of LLMs remains an open question. In the future, we will extend our approach for the models that use grouped-query attentions and will investigate the MoE models as well.

\section*{Ethics Statement}
We recruited annotators using Amazon Mechanical Turk for human evaluations. Because the LLM-generated summaries may contain inappropriate or sensitive language, we reviewed the summaries beforehand and found no problematic samples.


\appendix
\section{Dataset Details}
\label{appen:data} 
For the neural network regression, we used the dataset generated from each sequence of summaries, excluding input token prompts. We randomly divided the datasets into 90\% for training and 10\% for validation. Table~\ref{tab:data_detail} shows statistics.

\begin{table}[h]
    \centering
    \resizebox{0.5\textwidth}{!}{
    \begin{tabular}{cccc}
        \toprule
        \rowcolor{gray!10}
        \textbf{Model} & \multicolumn{1}{c}{\textbf{No-constraint}} & \multicolumn{1}{c}{\textbf{Length}} & \multicolumn{1}{c}{\textbf{Priming}} \\ 
        \midrule 
        Llama-2-7B-Chat & 21,385 & 24,121 & 25,394\\
        Llama-2-13B-Chat & 26,526 & 28,590 & 20,291\\
        Llama-2-13B-Chat(fine-tuned) & 14,826 & 16,373 & 14,884\\
        Llama-2-70B-Chat & 20,885 & 15,707 & 19,870\\
        Llama-3-8B-Instruct & 17,952 & 22,853 & 13,366\\
        Phi3-mini-4k-Instruct & 30,552 & 18,500 & 25,160\\
        Phi3-small-8k-Instruct & 25,578 & 30,938 & 18,841\\
        \bottomrule
    \end{tabular}
    }
    \caption{Dataset Statistics.}
    \label{tab:data_detail}
\end{table}

\section{Experimental Details}
\label{sec:appendixA}

\begin{table}[h]
    \centering
    \resizebox{0.3\textwidth}{!}{
    \begin{tabular}{cc}
        \toprule
        \rowcolor{gray!10}
        \textbf{Parameter} & \textbf{Value} \\
        \midrule
        Epochs & 1,000 \\
        Batch size &  32, 64\\
        Learning rate & 1e-3\\
        Dropout rate & 0.1\\
        Patience & 5, 10\\
        Loss & MSE\\
        Activation & ReLU\\
        \bottomrule
    \end{tabular}
    }
    \caption{Hyperparameters.}
    \label{tab:params}
\end{table}

\noindent \textbf{Computing Interfaces.}\label{appen:infra}
We used the following GPUs:
\begin{itemize}
    \item NVIDIA A100 GPU for Llama-2-70B-Chat
    \item NVIDIA A6000 GPU for other LLMs
\end{itemize}

\noindent \textbf{Hyperparameters.}
Table~\ref{tab:params} shows the hyperparameters used in our experiments.
For the neural network regression to predict the generation time step from all hidden unit (Table~\ref{tab:all_r2score}, Table~\ref{tab:all_r2score8bit}, and Figure~\ref{fig:fig2}), the batch size was set to 32 with an early stopping patience of 10 epochs. For the neural network regression to predict the generation time step from each individual hidden unit (Table~\ref{tab:eachhidden}, Figure~\ref{fig:zero} and Figure~\ref{fig:zero2}), the batch size was set to 64 with an early stopping patience of 5 epochs.

\section{Other Decoding Strategies}\label{appen:other_decoding}
\begin{figure*}[h]
\centering
    \includegraphics[width=1\textwidth]{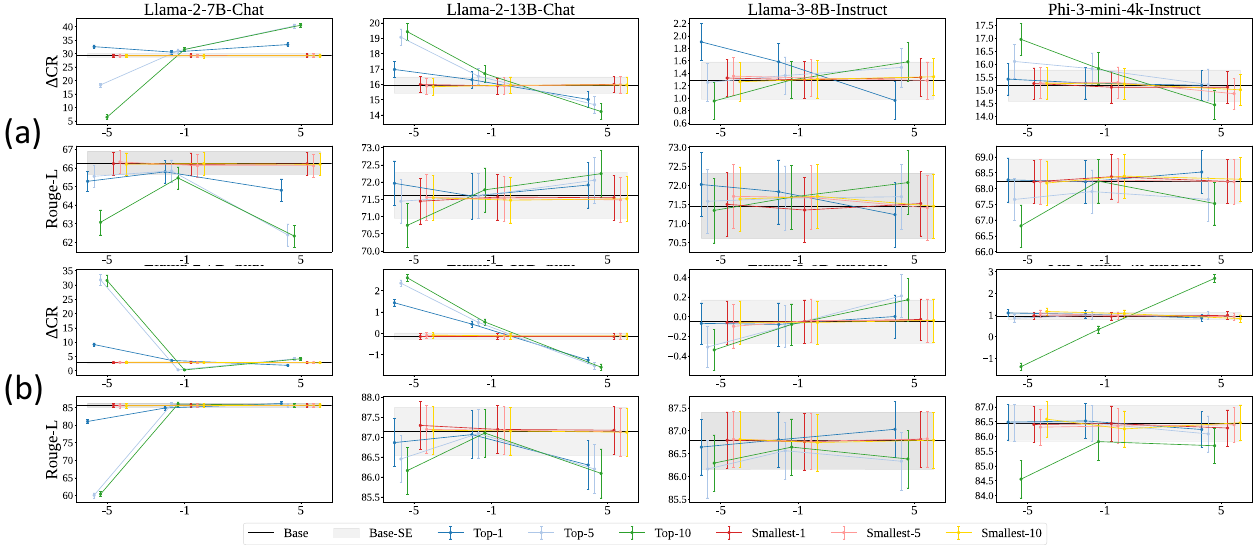}
    \caption{Results in (a) zero-shot settings and (b) fine-tuning settings using the Priming prompt with Beam decoding.}
    \label{fig:Beam}
\end{figure*}

\begin{figure*}[h]
\centering
    \includegraphics[width=1\textwidth]{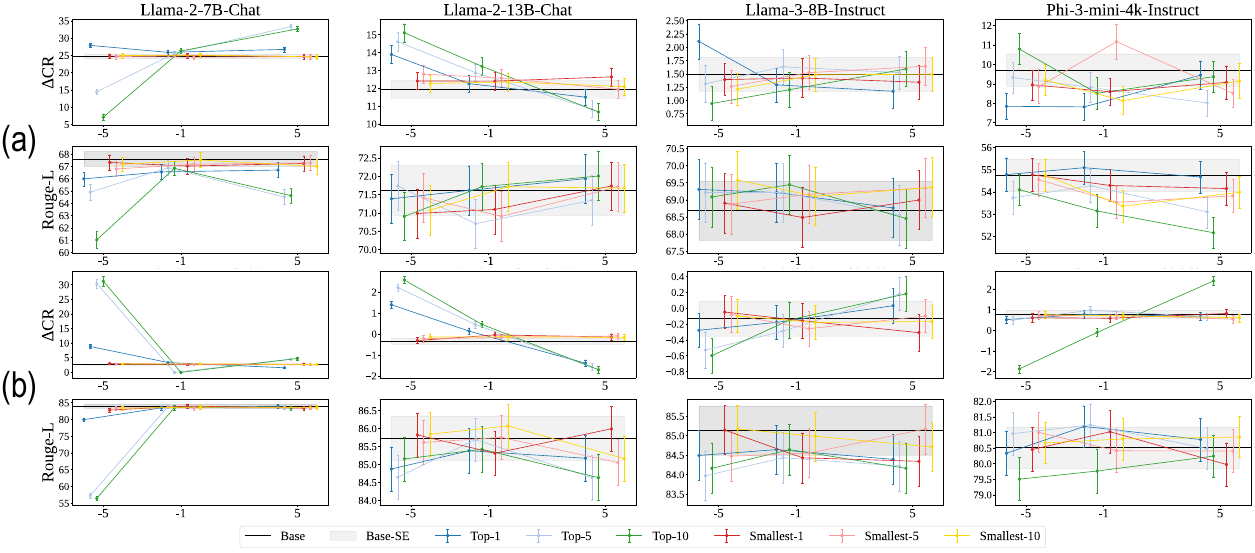}
    \caption{Results in (a) zero-shot settings and (b) fine-tuning settings using the Priming prompt with top-\textit{k} sampling.}
    \label{fig:Topk}
\end{figure*}

Figure~\ref{fig:Beam} and~\ref{fig:Topk} show additional experimental results for disentangling the top- and smallest-\textit{k} hidden units using beam and top-\textit{k} sampling methods. The beam size was set to three, and top-\textit{k} sampling was set to 10.
We observed consistent results with the greedy decoding method.

\section{Machine Translation and Story Generation}\label{appen:trans}
\begin{figure*}
\centering
    \includegraphics[width=0.9\textwidth]{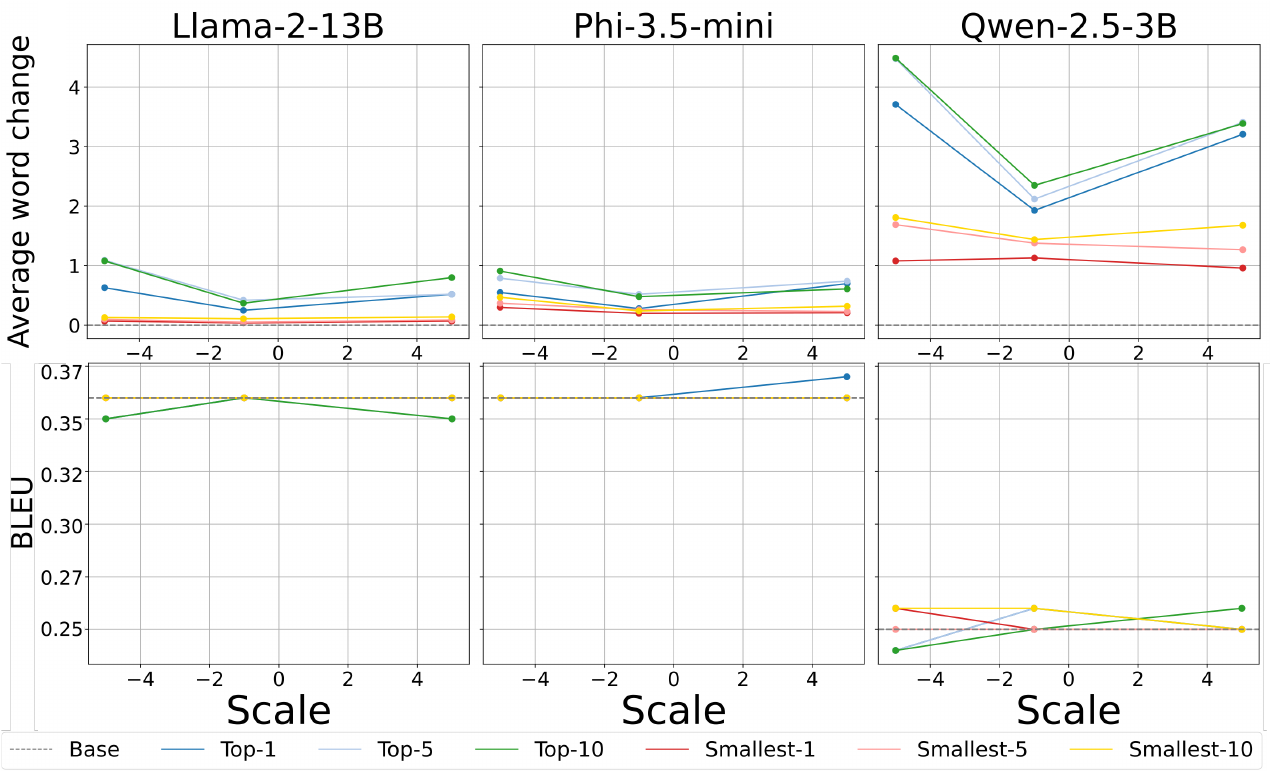}
    \caption{Experimental results on the WMT16 test dataset. We evaluated translation quality using the average of BLEU-1 and BLEU-2 scores.}
    \label{fig:Trans_result}
\end{figure*}

\begin{figure*}
\centering
    \includegraphics[width=0.9\textwidth]{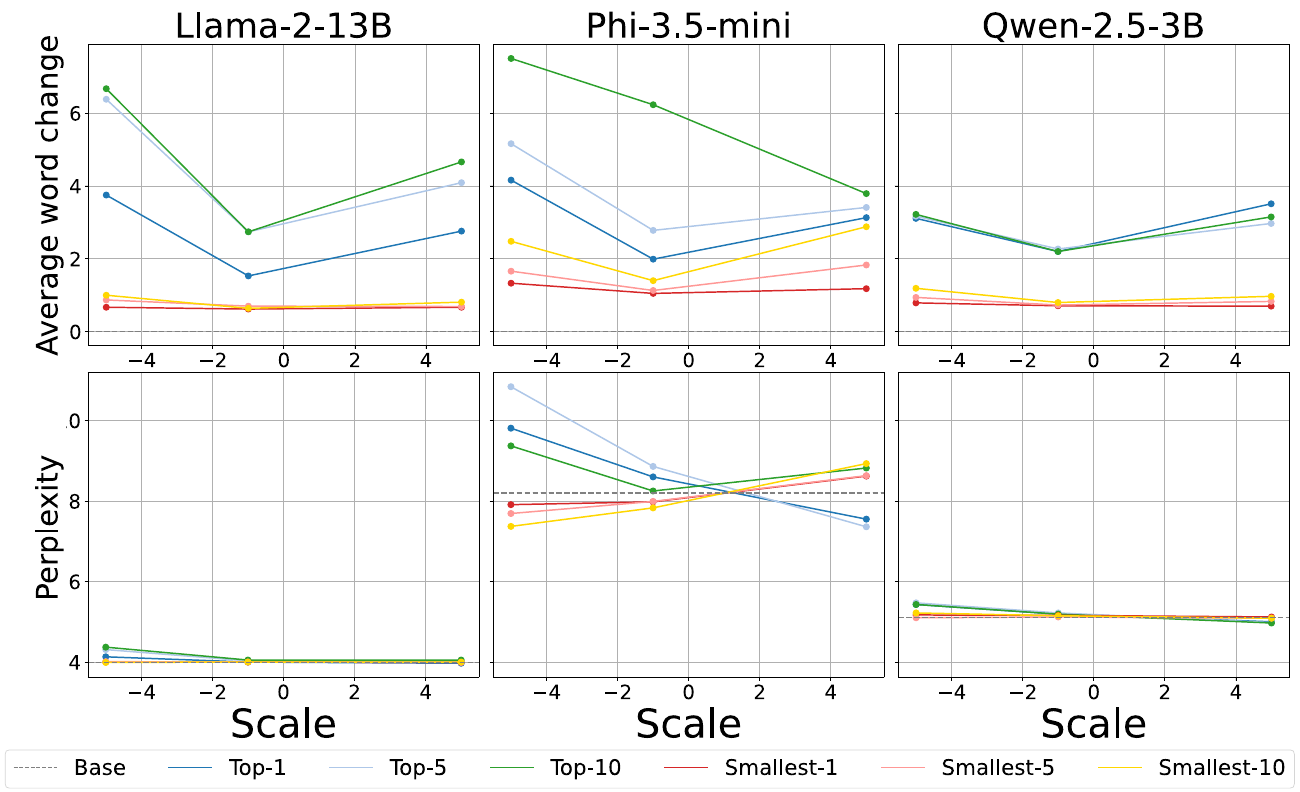}
    \caption{Experimental results on the ROCStories test dataset. We used the Qwen2.5-7B-Instruct model to evaluate perplexity (PPL). Lower PPL scores indicate better text quality.}
    \label{fig:roc_result}
\end{figure*}

We conducted additional experiments on two tasks: machine translation and story generation. For machine translation, we randomly sampled 500 instances from the WMT16 test dataset~\cite{bojar-etal-2016-findings}. For story generation, we used the ROCStories test dataset comprising 1,571 instances~\cite{mostafazadeh-etal-2016-corpus}. We evaluated these tasks in zero-shot settings, disentangling length-related units identified from summarization prompts. Figures~\ref{fig:Trans_result} and~\ref{fig:roc_result} show the results. Interestingly, we found that length-specific units are globally shared across tasks and can be adjusted without sacrificing informativeness.

\begin{tcolorbox}[colframe=black, colback=white, coltitle=black,  boxrule=0.5pt, title=\textbf{Machine translation prompt that considers \quotes{priming}}, colbacktitle=white]
\small
Sentence that consists of \{len(en)\} tokens:\\
 \{en\}\\
The sentence translated into German that consists of \{len(de)\} tokens with \{len(de)-len(en)\} additional tokens would be:\\
Sentence that consists of \{len(en)\} tokens:\\
{en}\\
The sentence translated into German that consists of \{len(de)\} tokens without \{len(en)-len(de)\} tokens would be:\\
\end{tcolorbox}

\begin{tcolorbox}[colframe=black, colback=white, coltitle=black,  boxrule=0.5pt, title=\textbf{Story generation Prompt}, colbacktitle=white]
\small
You are given the first four sentences of a short story. Please write a coherent fifth sentence that naturally concludes the story.\\
    Story:\\
    1) \{sentence 1\}\\
    2) \{sentence 2\}\\
    3) \{sentence 3\}\\
    4) \{sentence 4\}\\
    Now, write the fifth sentence:
\end{tcolorbox}

\section{Other case study}\label{appen:other}

\begin{table*}
    \begin{adjustbox}{width=1.9\columnwidth,center}
    \centering
    \begin{tabular}{p{0.2\columnwidth}p{0.2\columnwidth}p{2.0\columnwidth}p{0.2\columnwidth}}
        \toprule
        \rowcolor{gray!10}
        \multicolumn{2}{c}{\raisebox{-0.8\height}{\textbf{Type}}} & \raisebox{-0.8\height}{\textbf{Text}} & \textbf{Length (\#word)} \\
        \midrule
        \multicolumn{2}{c}{\multirow{3}{*}{Source}} & South African captain Graeme Smith hailed "an incredible win" for his team after they clinched an emphatic ten-wicket victory on the fifth day of the second and final Test against India at Kingsmead on Monday. & 35 \\
        \multicolumn{2}{c}{Gold}   & Graeme Smith hailed an incredible win. & 6 \\
        \midrule
        \multirow{1}{*}{\textbf{Top-10}}& Scale -10 & S. & 1 (-8) \\
        \midrule
        \multirow{1}{*}{\textbf{Base (Scale 1)}} &
         & South African captain Graeme Smith hailed an incredible win. & 9 \\
        \midrule
        \midrule
        \multicolumn{2}{c}{Source} &Unknown assailants blew up a natural gas pipeline in Egypt, a security source said. & 14 \\
        \multicolumn{2}{c}{Gold}   & Assailants blew up a gas pipeline in Egypt. & 8 \\
        \midrule
        \multirow{1}{*}{\textbf{Top-10}}& Scale -10 & AssBlewUpNatGasPipEgy. & 1 (-8) \\
        \midrule
        \multirow{1}{*}{\textbf{Base (Scale 1)}} &
         & Assailants blew up a natural gas pipeline in Egypt. & 9 \\
        \bottomrule
    \end{tabular}
    \end{adjustbox}
    \caption{Case studies by scaling factors using Llama-2-13B-Chat with zero-shot priming.}
    \label{tab:case2}
\end{table*}

Table~\ref{tab:case2} shows case studies.  
We found that the generated summaries ended abnormally early or that tokens were generated without spaces when extreme numeric values, such as -10, were used. This resulted in cases where the R-L scores significantly decreased with extreme scaling factors.

\end{document}